\documentclass[conference]{IEEEtran}
\IEEEoverridecommandlockouts
\usepackage{cite}
\usepackage{amsmath,amssymb,amsfonts}
\usepackage{algorithmic}
\usepackage{graphicx}
\usepackage{textcomp}
\usepackage{xcolor}
\def\BibTeX{{\rm B\kern-.05em{\sc i\kern-.025em b}\kern-.08em
    T\kern-.1667em\lower.7ex\hbox{E}\kern-.125emX}}
    
\newcommand{\bluHL}[1]{#1}
\newcommand{\eat}[1]{}
\usepackage{booktabs}
\def\authorrefmark#1{\raisebox{0pt}[0pt][0pt]{\textsuperscript{\footnotesize\ensuremath{\ifcase#1\or *\or \dagger\or \ddagger\or%
    \mathsection\or \mathparagraph\or \|\or **\or \dagger\dagger%
    \or \ddagger\ddagger \else\textsuperscript{\expandafter\romannumeral#1}\fi}}}}

\begin{document}

\title{When Single Event Upset Meets Deep Neural Networks: Observations, Explorations, and Remedies}

\author{\IEEEauthorblockN{Zheyu Yan\authorrefmark{1}, Yiyu Shi\authorrefmark{2}, Wang Liao\authorrefmark{3}, Masanori Hashimoto\authorrefmark{3}, Xichuan Zhou\authorrefmark{4}, Cheng Zhuo\authorrefmark{1}} \IEEEauthorblockA{\authorrefmark{1} Zhejiang University, \{yanzheyu, czhuo\}@zju.edu.cn  \authorrefmark{2} University of Notre Dame, yshi4@nd.edu\\ \authorrefmark{3} Osaka University, \{wang.liao,hasimoto\}@ist.osaka-u.ac.jp \authorrefmark{4} Chongqing University zxc@cqu.edu.cn}}

\eat{
\author{\IEEEauthorblockN{Zheyu Yan}
\IEEEauthorblockA{
\textit{Zhejiang University}\\
Hangzhou, China \\
yanzheyu@zju.edu.cn}
\and
\IEEEauthorblockN{Yiyu Shi}
\IEEEauthorblockA{
\textit{University of Notre Dame}\\
Notre Dame, US \\
yshi4@nd.edu}
\and
\IEEEauthorblockN{Wang Liao}
\IEEEauthorblockA{
\textit{Osaka University}\\
Osaka, Japan \\
wang.liao@ist.osaka-u.ac.jp}
\\
\and
\IEEEauthorblockN{Masanori Hashimoto}
\IEEEauthorblockA{
\textit{Osaka University}\\
Osaka, Japan \\
hasimoto@ist.osaka.ac.jp}
\and
\IEEEauthorblockN{Xichuan Zhou}
\IEEEauthorblockA{
\textit{Chongqing University}\\
Chongqing, China \\
zxc@cqu.edu.cn}
\and
\IEEEauthorblockN{Cheng Zhuo}
\IEEEauthorblockA{
\textit{Zhejiang University}\\
Hangzhou, China \\
czhuo@zju.edu.cn}
}
}

\maketitle

\begin{abstract}
Deep Neural Network has proved its potential in various perception tasks and hence become an appealing option for interpretation and data processing in security sensitive systems. However, security-sensitive systems demand not only high perception performance, but also design robustness under various circumstances. Unlike prior works that study network robustness from software level, we investigate from hardware perspective about the impact of Single Event Upset (SEU) induced parameter perturbation (SIPP) on neural networks. We systematically define the fault models of SEU and then provide the definition of sensitivity to SIPP as the robustness measure for the network. We are then able to analytically explore the weakness of a network and summarize the key findings for the impact of SIPP on different types of bits in a floating point parameter, layer-wise robustness within the same network and impact of network depth. Based on those findings, we propose two remedy solutions to protect DNNs from SIPPs, which can mitigate accuracy degradation from 28\% to 0.27\% for ResNet with merely 0.24-bit SRAM area overhead per parameter.
\end{abstract}


\section{Introduction}
 
Deep neural networks ($abbrev.$ DNNs) have recently attracted enormous attention due to the success in various perception tasks \cite{zheng2019learning, cheng2018intrinsic} and it is an appealing idea to adopt DNNs in security sensitive systems for in-depth inference and efficient data processing, such as autonomous automobile and medical monitoring. On the other hand, the robustness of DNN itself is of great concern for such security related applications and hence has been widely studied. Various methods, including adversarial example \cite{goodfellow2014explaining} and fault injection \cite{barenghi2012fault}, are devised to attack DNNs. Their aim is to fool the networks to generate adversarial outputs. 

Other than such carefully generated perturbations, adding small but random noise to DNNs may also induce severe damage. Stevenson \emph{et. al.}  and Cheney \emph{et. al.} \cite{stevenson1990sensitivity, cheney2017robustness} analyze the impact of random numerical noise to the weights of DNNs and observe significant degradation in classification accuracy when certain layers are "polluted". Although such noises are numerically small, they are not necessarily insignificant when implemented in hardware. 

\begin{figure}[ht]
  \centering\vspace{-0.5cm}
  \includegraphics[width=1\linewidth]{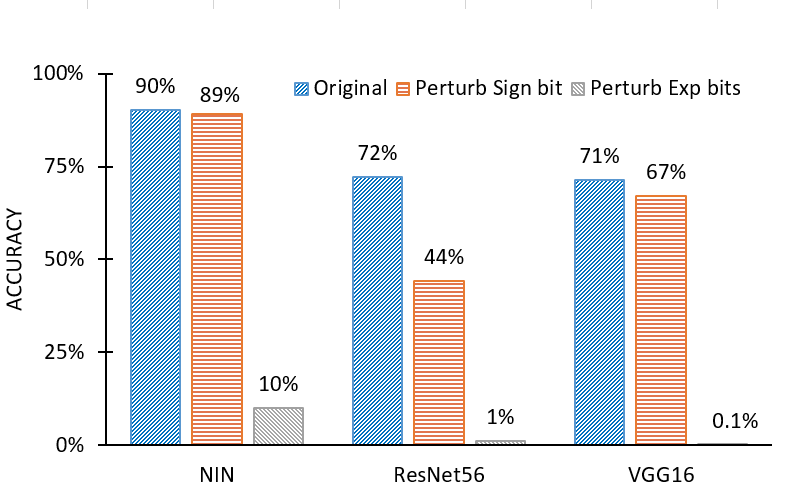}\vspace{-0.4cm}
  \caption{{Classification accuracy for different neural networks (NIN, VGG16 and ResNet56) under different types of SEU-induced perturbation. Blue: the accuracy for the original network; Orange: the minimum accuracy with perturbation on one weight's sign bit across all the weights; Gray: the minimum accuracy for perturbation on one exponent bit across all the exponent bits of the weights.}}
  \label{fig:Acc}\vspace{-0.3cm}
\end{figure}

\bluHL{Thus, Donato \emph{et al.}, Reagen \emph{et al.} and Sha \emph{et al.} \cite{donato2018chip, reagen2018ares, sha2018design} investigate the robustness of DNNs from emerging device, architecture and system perspective. However, they treated all error conditions homogeneously and sampled only a small portion of the errors to analyze the average effects. In security sensitive applications, the focus is more on the worst case instead of average performances. \cite{ziegler1979effect} Thus, in this work, we conduct more thorough experiments to explore the worst case.} To reduce the complexity of our searches, we place our focus on the minimum perturbation that can occur in a digital system, $i.e.$, single bit flip. Single bit flip can be triggered by Single Event Upset ($abbrev.$ SEU) in daily lives with ionizing particles hitting storage devices and logic units \cite{hirokawa2016multiple}. In reliability analysis, SEUs are often manifested as "bit flips" in which the value of a single bit is reversed from "0" to "1" or vice versa \cite{pettit2018detecting}. Fig.~\ref{fig:Acc} demonstrates the impact of SEU induced parameter perturbation on different network architectures including Network in Networks (NIN) \cite{lin2013network}, VGG16 \cite{simonyan2014very} and Residual Neural Networks (ResNet) \cite{he2016deep}. The figure compares the case without perturbation and the cases with perturbations occurring at sign bit or exponent bit. The accuracy of ResNet56 (trained on CIFAR-100 \cite{krizhevsky2009learning}) drops to almost 1\% for bit flip on the exponent. The figure simply presents that the smallest perturbation in DNN parameters may cause serious damage. Since SEUs are not uncommon in our everyday lives, the aforementioned perturbation is actually detrimental to the security-sensitive systems \cite{cheney2017robustness}. 

Thus, in this paper, we thoroughly investigate the problem of SEU-Induced Parameter Perturbation (\textit{abbrev.} SIPP) for DNNs as well as its remedy solutions. To the best of our knowledge, this is the first work that studies the worst case of SIPP for DNNs. The contributions of the work include:
\begin{itemize}
    \item \bluHL{We formally define the fault models to study SEU induced perturbation and propose an experimental flow to measure the network robustness sensitivity to SIPP. Several key observations are then summarized for ResNet56 with the proposed flow.}
    \item We analytically explore the impact of SIPP on parameters and the propagation of SIPP to the network output (Section 4.1 and 4.2). The analysis provides us in-depth understanding of how SIPP affects the system and  provides us guidelines to investigate the weakness of other DNNs.
    \item We then thoroughly investigate the robustness of three representative DNNs, NIN, VGG16 and ResNet56 (Section 4.3). After the investigation, three key findings which confirm our observations are presented.
    \item Based on the findings, we propose two simple yet efficient remedy solutions, triple modular redundancy (TMR) and error-correction code (ECC) to ensure complete protection from SIPP (Section 5). Design trade-off is then explored between protection overhead and design robustness for the two methods.
\end{itemize}
Experimental results show that without any protection, SIPP on a $Sign$ bit for ResNet56 may easily induce more than 28\% accuracy degradation. ECC based protection scheme can reduce such degradation to 0.27\% with SRAM overhead of merely 0.24 additional bit per parameter on average.  

\section{Background}
\subsection{{Single Event Upset}}
SEU is a transient information destruction on memory or logic elements caused by an energetic ionizing radiation. Ionized radiation particles may generate electron-hole pairs when they penetrate into the silicon substrate of a transistor \cite{pettit2018detecting}. After electron-hole pairs are generated, their transportation, such as diffusion and drift, collects electric charge to the drain region of the transistor. In memory elements, the collected charge accumulates and finally induces a glitch in the affected transistor, upsetting the stored information. In terrestrial environment, SEU is generally induced by two kinds of particles, alpha particles emitted from package material and neutrons originating from cosmic ray. In security sensitive applications, package material with low alpha particle emission is used to mitigate alpha particle induced SEU. However, since abundant neutrons in the cosmic ray go through materials on the ground and consequently they are difficult to eliminate with shielding, SEU threatens terrestrial VLSIs and hence demands targeted protections.

\subsection{Deep Neural Networks}
A typical feed forward DNN is a collection of convolution, activation, normalization, pooling and fully connected ($abbrev.$ FC) layers. The network is specified by a set of parameters, including weights, biases means and variances. Since means and variances can be written in the form of weights and biases, for simplicity, $parameters$ specified in this paper just refer to weights and biases.

A convolution layer applies 2D convolutions over an input signal composed of several input planes, $i.e.$, \emph{feature maps}. The output is then a 3D tensor with a similar shape. In the simplest case, the output value of the convolution layer with $C_{in}$ input channels and $C_{out}$ output channels can be precisely described as:
\begin{equation}\label{eq:Conv}
    \begin{split}
        O (q) = b (q) + \sum_{p=0}^{C_{in}-1} K(q, p) \otimes I(p), \ \ q=1,\dots,C_{out}
    \end{split}
\end{equation}
in which $I(\cdot)$ and $O(\cdot)$ denote the input and output planes, $\otimes$ is the convolution operation. $K$ is a set of 2D convolution kernels, with each corresponding to one pair of input and output planes. $b$ is a set of {scalars} globally added to each 2D output plane. The operation of FC layers can also be represented by Eq.~\eqref{eq:Conv} using scalars instead of 2D planes. The activation and pooling layers are normally hard-wired without extra parameters. The normalization layers normalizes each channel's output with trained means and variances.

\subsection{IEEE Standard for Floating Point Arithmetic in Hardware Implementation}

According to IEEE standard \cite{4610935}, 32-bit FP numbers commonly used in DNNs can be represented by:
\begin{equation}\label{eq:FP}
  x = (-1)^{Sign} \times (1 + Fraction) ^{Exponent - Bias}
\end{equation}
in which $Sign$ is the sign bit. $Exponent$ is an unsigned integer for exponent that uses the second to the ninth bits. In a 32-bit FP number, $Bias$ is 127. $Fraction$ is a fixed point number represented by the rest of the bits with the highest bit representing $2^{-1}$.

\section{Observations}\label{sect3}


\subsection{Single Event Upset in Neural Networks}
Hirokawa \emph{et al.} \cite{hirokawa2016multiple} show that, when exposed to terrestrial neutrons, a bit in SRAM has a probability of $1.33\times10^{-24}$ to flip in 1 $ns$. For all the parameters in one Neural Network, assuming that the flip of each bit is independent, the probability of at least one bit flip is:
\vspace{-0.1cm}
\begin{equation}\label{eq:Pos}
    \begin{split}
        P_{flip} & = 1 - ((1 - P_{single})^{N * W})^{T/t} \\
        & =\sum^{n-1}_{i=0} \binom{n}{i} (-1)^{n-i+1} P_{single}^{n-i}, n = N \times W \times T/t \\
        & \approx N \times W \times T/t \times P_{single} 
    \end{split}
\vspace{-0.1cm}\end{equation}
in which $N$ is the number of trained parameters, $W$ is the data width of each parameter, $T$ is device life time, $t$ is the test time interval, $i.e.$, 1 $ns$ as above, and $P_{single}$ is the probability for one bit flip within $t$. The approximation in Eq.~\eqref{eq:Pos} is appropriate when $N \times W \times T/t << 1/P_{single}$, which usually holds for DNNs \cite{hirokawa2016multiple}. For a typical DNN with more than 10M parameters, in one month, the probability of having at least one bit flip would be as high as 10\%, which is hazardous for the aforementioned security-sensitive scenarios and hence demands in-depth understanding of the impact of SEU induced perturbation.


\subsection{Fault Model}

It is crucial to select an appropriate fault model to measure the impact of SEU-induced failures to DNNs. In our experiments, for simplicity, we rely on the following assumptions for our fault model:
\begin{enumerate}
\item The training and inference processes are fault-free, and the fault is only introduced by hardware failures in storage devices, which preserves the network's parameters.
\item Among all the parameters, only one bit of one parameter is with fault and the others are fault-free.

\end{enumerate}

\subsection{SSIPP: Measuring Neural Network Robustness}

\bluHL{To analyze the effect of SIPP,} it is crucial to have a rigorously-defined metric to measure the impact of SIPP on DNNs. Apparently the most straightforward way is to compare the output difference for the same set of inputs. Thus, here we propose the concept of \textit{Sensitivity to SIPP} ($abbrev.$ SSIPP) as a quantitative measure to assess the differences of network robustness. The definition of SSIPP is as below:\\
\textbf{Definition:} For SIPP on a particular bit $i$, its performance change due to the perturbation from the original network can be calculated by $\Delta P_i = P_{original} - P_{SIPP_i}$, where $P(\cdot)$ denotes the performance measure for a network.\footnote{Performance measure is specific to the network type and application, $e.g.$, cross-entropy for segmentation, accuracy for classification.} Then the robustness measure of $SSIPP$ is defined as:
\begin{equation}
    SSIPP=\max_{\forall\,i}\Delta P_i
\end{equation}
\bluHL{Unlike similar performance degradation based methods\cite{reagen2018ares,donato2018chip} have been proposed, SSIPP focus on searching all possible fault patterns (flipping all the bits of a network) and finding the worst case.} If the perturbed bits are limited to one particular layer, then it measures SSIPP of this particular layer for the network. The one with a smaller SSIPP is considered more robust.

\begin{figure}[ht]
  \centering\vspace{-0.3cm}
  \includegraphics[width=1\linewidth]{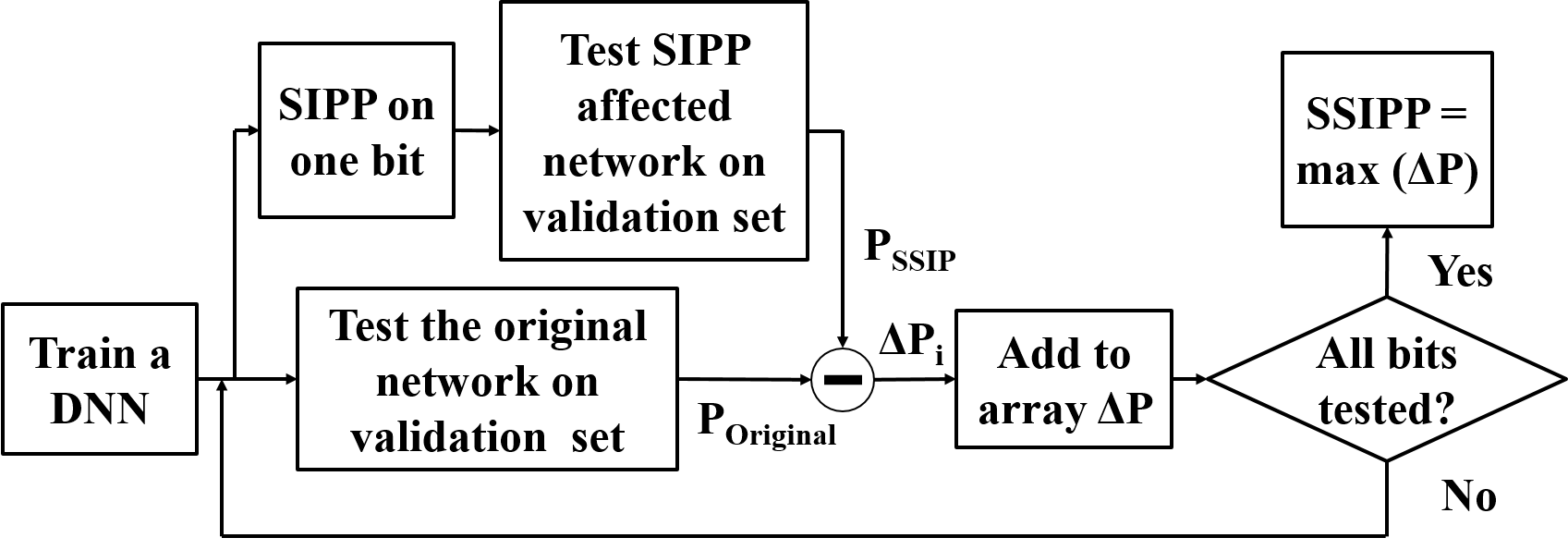}\vspace{-0.3cm}
  \caption{Flow to measure SSIPP for a DNN.}
  \label{fig:AccDWF}\vspace{-0.3cm}
\end{figure}

The work flow to measure SSIPP is illustrated in Fig.~\ref{fig:AccDWF}. First, the original network is tested on the validation set to obtain the reference performance $P_{original}$. Then, we induce SIPP on one bit of the network to obtain a perturbed network, which is also tested on the same validation set to measure its performance $P_{SIPP}$. 
After that we can compute the performance change $\Delta P_{i}$ due to this particular bit. Finally, by visiting every bit of the network, we obtain the network robustness measure $SSIPP$. 

\subsection{Observations}

With the given definition of SSIPP and flow in Fig.~\ref{fig:AccDWF}, we conduct a preliminary experiment to understand the impact of different SIPP patterns. We use ResNet-56 as the underlying DNN. The perturbation is introduced to the MSBs of a FP number,$i.e.$ sign bit, the highest exponent bit and the highest fraction bit of the parameters, respectively. After checking all the parameters in the network, Table~\ref{tab:RESB} summarizes SSIPP for different layers of ResNet-56. From the table, there are a few key observations that can be made for the impact of SIPP on ResNet56:
\begin{itemize}
    \item {\textbf{Observation 1:}} the highest exponent bit consistently has the highest SSIPP across different layers while the impact of SIPP on a fraction bit is very limited.
    \item {\textbf{Observation 2:}} the first layer, which directly deals with the input stream, has higher SSIPP than the other layers, then follow the lasted layers. This indicates higher robustness for the hidden layers in the middle of a network.

\end{itemize}
Then a natural question for the observations is:``Are those observations universal or unique to ResNet56?'' This motivates us to conduct more in-depth explorations to understand the DNN robustness to SIPP, which will be discussed in the next section. 

\vspace{0cm}
\begin{table}[ht]
  \caption{SSIPP for different layers in ResNet56: input, residual (stack 1, 2 and 3) and FC.}
  \label{tab:RESB}
  \begin{tabular}{c|ccccc}
    \toprule
    $SSIPP$  & Input & Stack 1 & Stack 2 & Stack 3 & FC\\
    \midrule
     Sign&28.09\%&6.43\%&2.08\%&0.27\%&0.90\%\\
     Ex1&70.19\%&70.19\%&70.19\%&70.19\%&70.19\%\\
     Frac1&0.43\%&0.29\%&0.41\%&0.25\%&0.30\%\\
  \bottomrule
\end{tabular}
\end{table}
\vspace{-0.4cm}

\section{Explorations}
In this section, we conduct systematical analysis of the impact of SIPP on parameter value in a DNN and how SIPP propagates to the output. The analysis then provides theoretical explanations to the observations in the last section. 
\subsection{Understanding the Impact of SIPP on Parameters}
As is discussed in Sect.~\ref{sect3}, there are different types of bits in a 32-bit FP parameter, including $Sign$, $Exponent$ and $Fraction$ bits. Thus, SIPP on different bits may impose different impacts on the parameter value. For a pre-trained DNN with known parameters, we can analyze the relative error $\Delta_{rel}$ imposed by SIPP.
SIPP on $Sign$ bit will cause the parameter to take its opposite value, $i.e.$, $\Delta_{rel}=2$. For SIPP on $Exponent$ bit, flip patterns affect $\Delta_{rel}$ greatly. A "1" to "0" flip reduces the absolute value of the parameter closer to 0, resulting in a $\Delta_{rel}$ between 0.5 and 1.0 while a "0" to "1" flip multiplies the parameter by a power of two. $\Delta_{rel}$ for such a change is larger than 1 and up to $2^{128}$, which can be detrimental to the system. For $Fraction$ bit, SIPP only induces a relatively smaller change with $\Delta_{rel}$ between 0 and 0.5.

 Thus, this explains the first observation in the last section why we observe much higher SSIPP for $Sign$ and $Exponent$ bits than $Fraction$ bits.

\subsection{Understanding the Propagation of SIPP}

For different SIPP patterns, the impact of SIPP on $Exponent$ bit and  $Fraction$ bit are detrimental or very limited, respectively, as discussed in the last sub-section. The impact of SIPP on $Sign$ bit is uncertain, which needs thorough analysis. Moreover, analyzing the impact of SIPP on $Fraction$ and $Sign$ bits can also provide some insights for widely used fixed-point accelerators \cite{jiang2019achieving, zhang2015optimizing}. \eat{\bluHL{Moreover, an increasing number of DNN accelerators are using fixed-point components instead of the complex floating-point calculation. Since $Sign$ bit is also a critical bit in fixed-point numbers, analyzing the effect of SIPP on $Sign$ bits could also offer some insight for fixed-point accelerators.} }

Thus, in this section we will focus on analyzing the impact of SIPP on $Sign$ bit. Since SIPP on a $Sign$ bit in former layers of DNNs actually needs to go through multiple layers to reach output, it is crucial to understand how the perturbation is propagated and why it is not cancelled out during the procedure. In the following we will provide detailed analysis to understand the propagation procedure of SIPP for both FC and convolution layers. 

For the $l^{th}$ FC layer with $N$ inputs and $M$ outputs, $N\times M$ \emph{weights} and $M$ \emph{biases} are needed. The output of this FC layer is:
 \vspace{-0.1cm}
\begin{equation}\label{eq:original}
  O_{lj} = \sum^{N}_{i=1} I_{i}\times \omega_{lij} + b_{lj},\quad j = 1, 2, ..., M
 \vspace{-0.1cm}\end{equation}
where $\omega_{lij}$ and $b_{lj}$ are the corresponding weight and bias.

\begin{figure}[ht]
  \centering\vspace{-0.3cm}
  \includegraphics[width=1\linewidth]{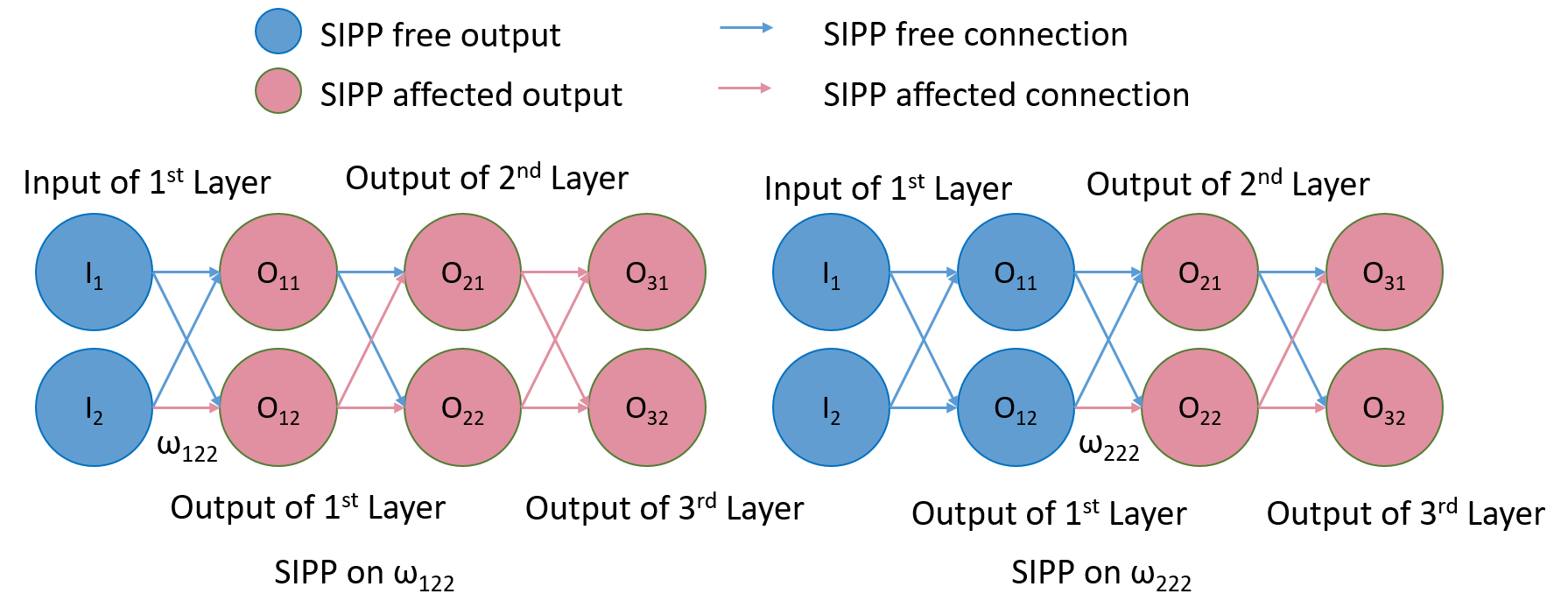}\vspace{-0.4cm}
  \caption{An example of perturbation propagation for FC layer: perturbation on $w_{122}$ at the 1st layer (left) and perturbation on $w_{222}$ at the second layer (right).}\vspace{-0.3cm}
  \label{fig:FCE}
\end{figure}

For simplicity we will use a portion of FC layers as shown in Fig.~\ref{fig:FCE} to demonstrate how the perturbation is propagated. The example contains 3 FC layers with 2 inputs and 2 outputs. The blue represents SIPP free data and connection, while the red represents SIPP affected data and connection. If ignoring nonlinear activation between FC layers for the moment, we can write down the output as the following:
 \begin{equation}\small{
\begin{split}\label{eq_full_begin}
 O_{3j} = &\omega_{31j} [\omega_{211}(\omega_{111} I_1 + \omega_{121} I_{2}) + \omega_{221}(\omega_{112} I_{1} + \omega_{122} I_{2})]\\
         +&\omega_{32j} [\omega_{212}(\omega_{111} I_1 + \omega_{121} I_{2}) + \omega_{222}(\omega_{112} I_{1} + \omega_{122} I_{2})]
\end{split}}
\end{equation}
where $j=1, 2$. Then for perturbation on the first and second layers, for example, $\omega_{122}$ and $\omega_{222}$, the changes at the outputs from the original values ($\Delta_{3j}=O_{3j}^{'}-O_{3j}$) are:
\begin{align}\label{eq:change1}
    \Delta_{3j}= 2\cdot(\omega_{31j}\omega_{221}+ \omega_{32j} \omega_{222}) \omega_{122} I_2, \ \ & for\ \omega_{122}\\\label{eq:change2}
   \Delta_{3j} = 2\cdot\omega_{32j}\omega_{222}(\omega_{112} I_1 + \omega_{122} I_2),\ \ & for\ \omega_{222}
\end{align}

From a layer-wise perspective when evaluating SIPP for all the parameters in a particular layer, we may reasonably assume the weights from the same layer are similar (but can be significantly different from layer to layer). For $I_1$ and $I_2$ from the same input source or distribution, so as those weights in the same layers. We then may claim that the two perturbations impose very similar impacts on the output, $i.e.$ $\Delta_{SIPP}$ is statistically insignificant. In other words, SIPP on sign bit from different layers may eventually result in very similar impact on the final output of the network if the network only contains linear operations as in Eq.~\eqref{eq_full_begin}-\eqref{eq:change2}.

Similar analysis can be conducted on the convolution layer to find the impact of SIPP on the sign bit of weight and bias using the following formulations:
\begin{equation}\label{eq:convlution}
 OF_{lq}^{'} = OF_{lq} - 2 \times\omega_{lpqrs} \times IF_{lp} \otimes S_{lrs}
\end{equation}
\begin{equation}\label{eq:CF_error}
 OF_{lq}^{'} = OF_{lq} - 2 \times b_{lq}
\end{equation}
where $l$ is the layer index, $p$ and $q$ are the input and output feature map indexes, $r$ and $s$ specify the weight location for the convolution kernel, $\omega_{lpqrs}$ is the perturbed weight. $OF_{lq}^{'}$ is the perturbed $2D$ output feature map, while $IF_{lp}$ and $OF_{lq}$ are the original input and output, respectively. In Eq.~\eqref{eq:convlution}, $S_{lrs}$ is a convolution kernel with the same size as the convolution kernel, with 1 at the perturbed weight's position and 0 for the others. 

\begin{figure}[ht]
  \centering\vspace{-0.3cm}
  \includegraphics[width=1.0\linewidth]{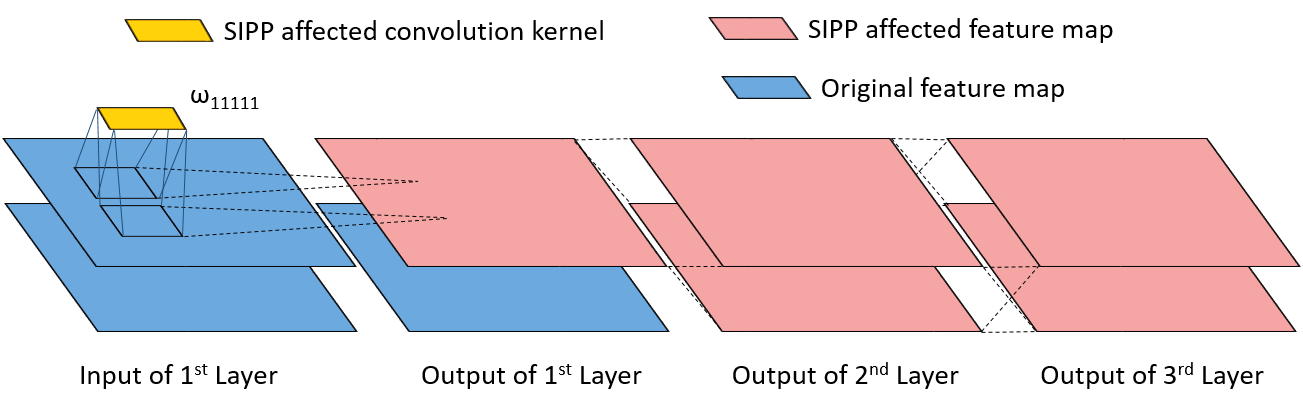}\vspace{-0.4cm}
  \caption{An example of perturbation propagation for convolution layer.}
  \label{fig:FMS}
\end{figure}

Unlike FC layer where the perturbation only impacts the connected data, the perturbation in convolution layer gets amplified and effectively propagates to a broader region through convolution of feature maps, as shown in Fig.~\ref{fig:FMS}. The affected region grows from a kernel to one feature map and then multiple feature maps. For perturbations on weights in two different layers, for example, $\omega_{11111}$ and $\omega_{21111}$, the output differences $\Delta_{3122}$ of (2,2) in the output feature map $OF_{31}$ for the two perturbations are:
\begin{equation}\small{
    \Delta_{3122(1)} = 
    2\sum_{q,r,s,i,j} \omega_{31qrs} \omega_{21qij}\omega_{11111}I_{11(r+i-3)(s+j-3)}}
\end{equation}
\begin{equation}\small{
    \Delta_{3122(2)} =  
    2\sum_{p,r,s,i,j}\omega_{3p1rs} \omega_{21111}\omega_{1p1ij} I_{1p(r+i-2)(s+j-2)}}\label{eq:ConvE}
\end{equation}
where $i$ and $j$ are the indexes to compute the $2D$ convolution, $\sum_{q,r,s,i,j}$ denotes the summation over each variable. From Eq.~\eqref{eq:ConvE}, we may draw a similar conclusion as the FC layers that, without activation layers, the effect of SIPP on weights from different convolution layers are almost equivalent for the output.

\bluHL{The analysis above shows that the key contributors of the layer-wise difference of SSIPPs are the activation layers. Thus, the analysis of activation layers is needed. The models analyzed in this work use ReLU layers for activation. The calculation of ReLU layers is:}
 $Out = max(0,Input)$,
\bluHL{which means $Inputs$ greater than 0 could propagate through while the smaller ones would be deactivated. As shown in Fig.~\ref{fig:FCE} and Fig.~\ref{fig:FMS}, the effects of SIPP on the former layers spreads wider than latter ones, they are less likely to be totally deactivated by activation layers, and thus more likely to propagate to the output and affect the classification results.}

The analysis above provides systematical support for observation 2 in the last section.

\subsection{Design Explorations}


\begin{table}[ht]
  \caption{Number of parameters (weight and bias) for different layers in the three DNNs: NIN, ResNet56 and VGG16.}
  \label{tab:size}
  \begin{tabular}{c|rr|rr|rr}
    \toprule
    &\multicolumn{2}{c|}{NIN}&\multicolumn{2}{c|}{ResNet56}&\multicolumn{2}{c}{VGG16}\\
    {Layer} &  \#Weight & \#Bias & \#Weight & \#Bias & \#Weight & \#Bias\\
    \midrule
     1&14k &192&0.4k&16 &1.7k &64\\
     2&30k &160&1.5k&16 &37k  &64\\
     3&15k &96 &3k  &32 &74k  &128\\
     4&460k&192&6.1k&64 &295k &128\\
     5&36k &192&-   &-  &590k &256\\
  Last&1.9k&10 &6.4k&100&4096k&1000\\
  \bottomrule
\end{tabular}
\end{table}

\begin{table*}[ht]
  \caption{SSIPP of different layers for the three neural networks with SIPP on the $Sign$ (Sign), $Exponent$ (Ex) and $Fraction$ (Fr) bits of weights (W) and biases (B).}
  \label{tab:AccWBE}
  \begin{tabular}{c|cc|cc|cc|cc|cc|cc|cc|cc|cc}
    \toprule
    &\multicolumn{2}{c|}{NIN/Sign}&\multicolumn{2}{c|}{NIN/Ex}&\multicolumn{2}{c|}{NIN/Fr}&\multicolumn{2}{c|}{Res56/Sign}&\multicolumn{2}{c|}{Res56/Ex}&\multicolumn{2}{c|}{Res56/Fr}&\multicolumn{2}{c}{VGG16/Sign}&\multicolumn{2}{c}{VGG16/Ex}&\multicolumn{2}{c}{VGG16/Fr}\\
    Layer&W & B & W & B & W & B & W & B & W & B & W & B & W & B & W & B & W & B\\
    \midrule
     1&1.0\%&2.2\%&80\%&80\%&0.2\%&0.4\%&28.1\%&16.9\%&70\%&70\% &0.4\%&0.4\%&4.2\% &2.4\% &70\%&70\%&0.9\%&0.9\%\\
     2&0.2\%&4.4\%&80\%&80\%&0.1\%&0.2\%&6.4\%&1.2\%  &70\%&70\% &0.3\%&0.2\%&0.1\% &0.4\% &70\%&70\%&0.1\%&0.2\%\\
     3&0.3\%&2.4\%&80\%&80\%&0.2\%&0.1\%&2.1\%&1.2\%  &70\%&70\% &0.4\%&0.5\%&0.1\% &0.1\% &70\%&70\%&0.1\%&0.1\%\\
     4&0.1\%&0.1\%&80\%&80\%&0.1\%&0.1\%&0.3\%&1.1\%  &70\%&70\% &0.3\%&0.4\%&0.1\% &0.1\% &70\%&70\%&0.1\%&0.1\%\\
     5&0.3\%&0.2\%&80\%&80\%&0.2\%&0.1\%&-    &-      &-   &-    &-    &-    &0.1\% &0.1\% &70\%&70\%&0.1\%&0.1\%\\
  Last&2.4\%&0.2\%&80\%&80\%&0.7\%&0.1\%&1.0\%&0.1\%  &70\%&70\% &0.3\%&0.1\%&0.0\% &0.0\% &70\%&70\%&0.1\%&0.1\%\\
  \bottomrule
\end{tabular}
\end{table*}
In this subsection we validate the impact of SIPP and our findings on different DNN architectures for image classification, the performance of which is measured by accuracy. Three representative architectures are employed in our design explorations, including a Network in Network (NIN) model (trained on CIFAR-10) \cite{lin2013network}, a 56-layer Deep Residual Network (ResNet56) (trained on CIFAR-100) \cite{he2016deep}, and VGG16 (trained on ImageNet) \cite{simonyan2014very,deng2009imagenet}. The three network architectures are selected to represent different types of feed forward deep neural networks. NIN is a model with no FC layer and hence helps us understand the role of convolution layers when evaluating network robustness. \bluHL{ResNet56 belongs to ResNets, a group of very deep neural network that consists of tens to hundreds of convolution layers.} Unlike ResNet56 with residual blocks, VGG net is a more classical deep convolutional neural network with simpler architecture and widely adopted by a variety of perception works \cite{ma2015hierarchical, he2017neural} and analyzed by a variety of accelerators \cite{jiang2019accuracy, zhang2015optimizing}. The numbers of parameters for the three neural networks are summarized in Table~\ref{tab:size}. Due to the size of DNN, we only present the results of the first five layers and the last layer of each DNN. 

Table~\ref{tab:AccWBE} demonstrates the impact of SIPP on the different types of bits, $i.e.$, $Sign$ (denoted by Sign), $Exponent$ (denoted by Ex) and $Fraction$ (denoted by Fr), of both weights (denoted by W) and biases (denoted by B). For each network, we calculate its SSIPP for a particular layer for the first five layers (denoted by layer 1 to 5) plus the last layer (denoted by Last). It is found that the impact of exponent bit is very prominent while fraction bit has very limited SSIPP. Moreover, sign bits of the first few layers have larger impacts than the rest layers, which is consistent with our analysis in the last subsection. \bluHL{The SSIPP on $Exponent$ bits in each layer is similar because they could always destroy the whole network into a random guesser.}

\begin{figure}[ht]
  \centering\vspace{-0.3cm}
  \includegraphics[width=1.0\linewidth]{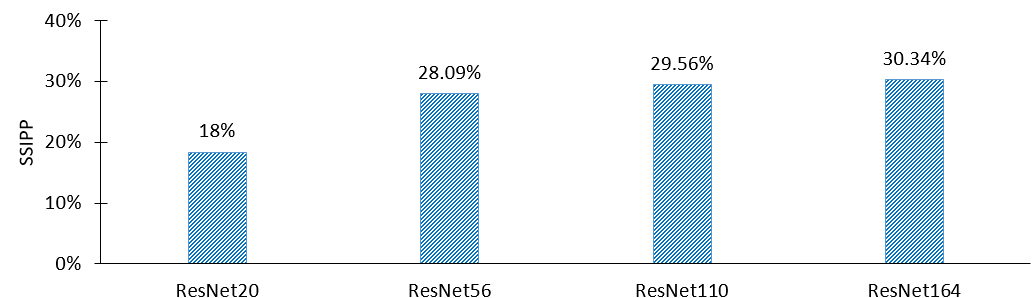}\vspace{-0.3cm}
  \caption{SSIPP of ResNets with different depth for perturbation on $Sign$ bit of the first layer.}\vspace{-0.3cm}
  \label{fig:resAccL}
\end{figure}

We further investigate the propagation of SIPP on the same network architecture but with different complexity. Fig.~\ref{fig:resAccL} compares SSIPP of first layer on $Sign$ bit for 4 residual networks with different depth, $i.e.$, ResNets with 20, 56, 110 and 164 layers. As shown in the Figure, with deeper network, the impact of SIPP is more prominent, which is also consistent with the findings in the last subsection. 

Thus, based on the observation, analysis and experimental results, we may summarize the following findings:
\begin{itemize}
    \item \textbf{Finding 1:} For the three types of bits in a 32-bit FP parameter, SIPP on $Exponent$ bit has the largest impact on network output while the impact from $Fraction$ bit is the minimal, with the impact of $Sign$ bit in the middle.
    \item \textbf{Finding 2:} SIPP on $Sign$ bit has layer-wise impact within the same network. The layer farther from the output is typically more sensitive and brings more significant change to the output.
    \item \textbf{Finding 3:} For two networks, with width of the network being the same, the deeper the network, the more sensitive the network is to SIPP.
\end{itemize}

\section{Remedies}
With the findings in the last section, the weakness of a network is more observable. Thus, it is necessary to further investigate the possible remedy solutions to the weakness. This section discusses two simple yet efficient remedies to the issues caused by SIPP and then investigates design trade-offs for the two methods.
\subsection{Triple Modular Redundancy for Parameter Protection}
For the parameters susceptible to SIPP, a natural solution is to provide multiple copies for the parameters of interest. Triple Modular Redundancy (TMR) \cite{lyons1962use} is just such a method that copies twice the circuit to be protected, thereby forming a group of three identical circuits with three outputs. The three outputs then go through a majority-voter to mask the fault and decide a single output. \bluHL{With the findings presented in previous sections,} we can prepare three identical copies of the parameters in SRAM to fully prevent SIPP. When parameters are fetched and sent to neural networks, the three copies of one parameter will go through a simple TMR circuit to chooses the correct output. 

\subsection{Error-Correcting Code}
Apparently, the area overhead of TMR based parameter protection can be significant. To resolve this issue, we further adopt error-correcting code (ECC) to protect parameters in SRAM with much smaller area overhead.

\begin{figure}[ht]
  \centering\vspace{-0.3cm}
  \includegraphics[width=1.0\linewidth]{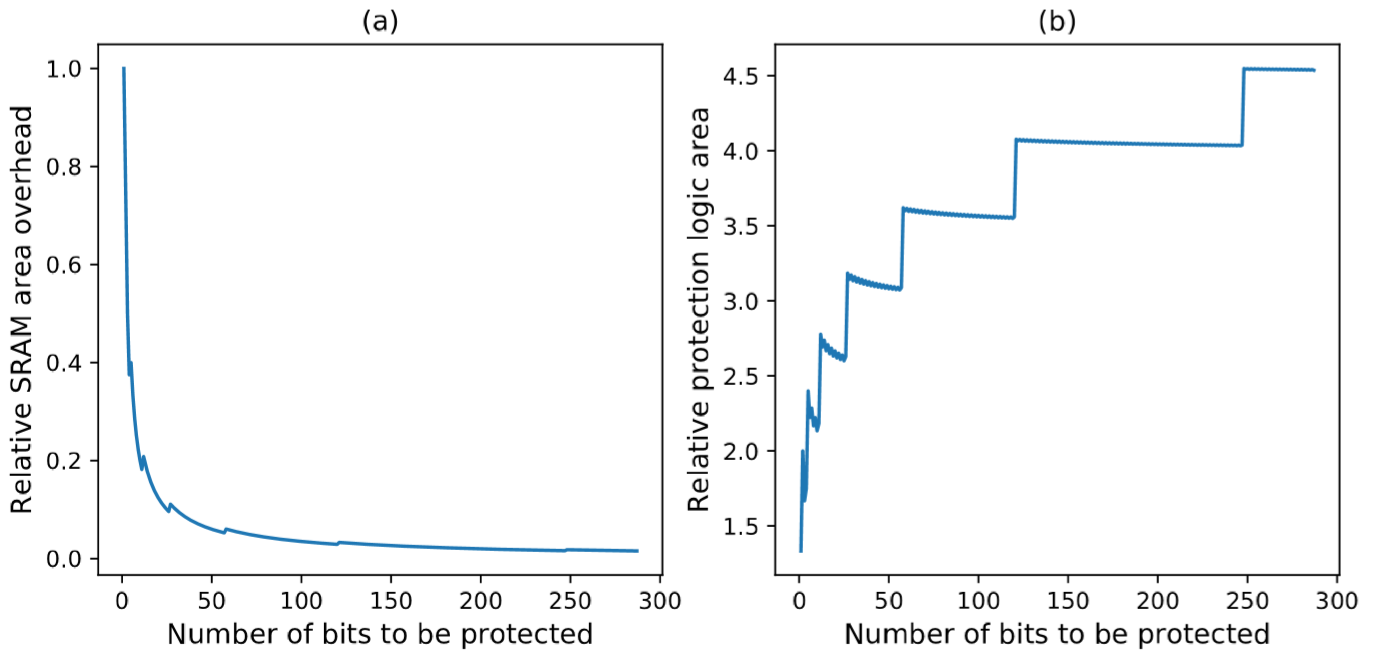}\vspace{-0.3cm}
  \caption{Relative SRAM (a) and logic (b) area overhead for ECC $w.r.t.$ TMR based protection.}
  \label{fig:TVEML}
\end{figure}

Hamming code is a family of binary linear ECCs by offering redundant correcting bits. With its single-bit protection feature, we can rely on Hamming code based ECC to fully protect the issues caused by SIPP. The number of redundant bits ($r$) to protect $d$ data bits needs to satisfy the following constraint: 
\vspace{-0.1cm}
\begin{equation} \label{eq:ECC}
    r + d \leq 2^r - 1
\vspace{-0.1cm}\end{equation}
Thus, the area overhead of ECC is exponentially smaller than TMR based approach. However, the SRAM area saving of ECC is at the cost of more complex logic to implement the protection and correction circuit. Fig.~\ref{fig:TVEML} presents the SRAM area and protection logic  overhead for ECC $w.r.t.$ TMR-based protection. It is found that, to protect 100 bits, with the SRAM and logic area overhead of the TMR-based method, only 3.5\% of the area overhead could be achieved at a cost of approximately 3.5$\times$ of protection logic.

\subsection{Design Trade-Off}

\begin{figure}[ht]
  \centering\vspace{-0.2 cm}
  \includegraphics[width=1\linewidth]{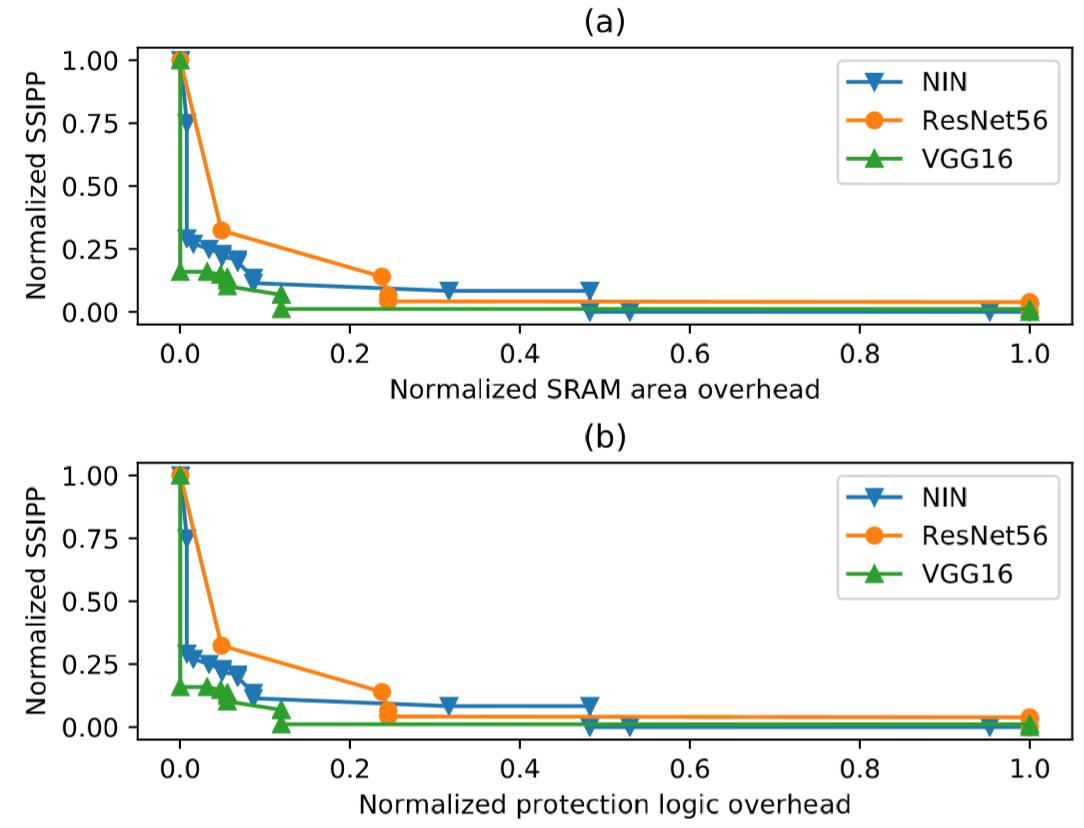}\vspace{-0.4cm}
  \caption{Normalized area overhead (normalized to full protection) using TMR based protection for three DNNs: (a) SRAM area overhead and (b) Protection logic area overhead for SSIPP (normalized to the case without protection).}\vspace{-0.5cm}
  \label{fig:TMRML}
\end{figure}

\begin{figure}[ht]
  \centering\vspace{0 cm}
  \includegraphics[width=1\linewidth]{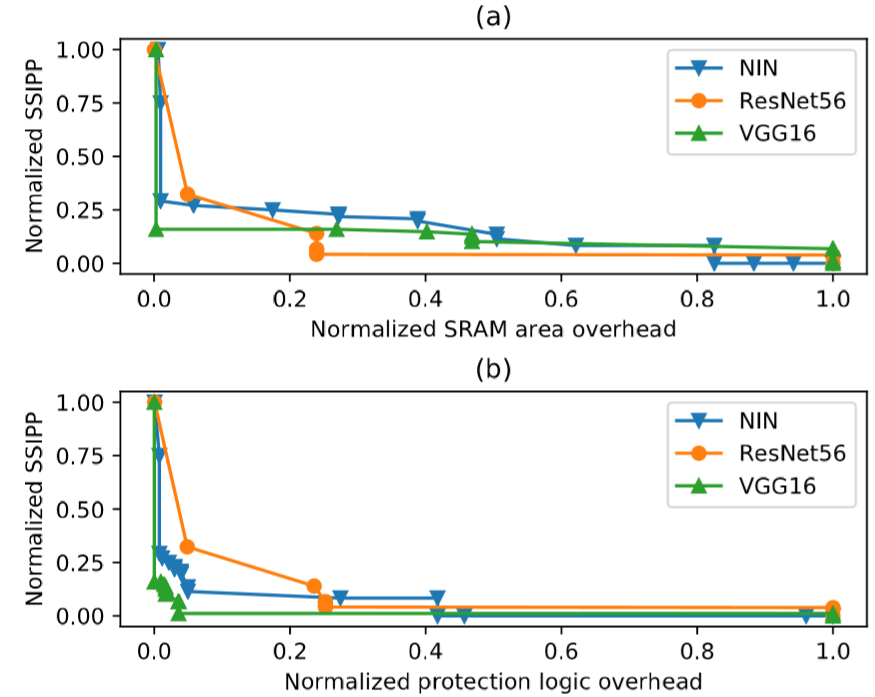}\vspace{0 cm}
  \caption{Normalized area overhead (normalized to full protection) using ECC based protection for three DNNs: (a) SRAM area overhead and (b) Protection logic area overhead for SSIPP (normalized to the case without protection).}
  \label{fig:ECCML}\vspace{0 cm}
\end{figure}

With the explorations in the last section, we are aware of the sensitivity of layers and bits for a DNN. Thus, instead of fully protecting all the parameters, we can further reduce both SRAM and protection logic area overhead by tolerating SIPP in non-sensitive bits. Fig.~\ref{fig:TMRML} demonstrates the area overhead for SRAM and protection logics, respectively, when implementing the TMR based protection. The area overhead is normalized to the case of full protection, $i.e.$, all the parameters are protected. The SSIPP on the y-axis is normalized to the case without any protection, $i.e.$, the wrost case SSIPP. The figure then provides design trade-off opportunities for the three DNNs. It can be seen that with merely 24\% SRAM and protection logic overhead, we can reduce the SSIPP for ResNet56 to only 2.08\%. Similar design trade-off can be conducted for ECC based protection, as shown in Fig.~\ref{fig:ECCML}(a) and (b). With 23\% SRAM area overhead and 25\% protection logic overhead, ECC based protection can reduce the SSIPP for ResNet56 to 2.08\%. By further increasing SRAM overhead to 25\%, SSIPP can be further reduced to 0.27\%.

\section{Conclusions}
In this paper, we investigate the robustness of DNNs from a hardware prospective about the impact of SIPP. We systematically define the fault models of SEU and then provide the definition of SSIPP as the robustness measure for the network. We then analytically explore the weakness of a network and summarize the key findings for impacts of SIPP on different types of bits in an FP parameter, layer-wise robustness within the same network and impact of network depth. Based on these findings, two remedy solutions can be adopted to protect DNNs from SIPP.

%
\bibliographystyle{ieeetr}
\bibliography{M8_References}

\eat{

\vspace{12pt}
\color{red}
IEEE conference templates contain guidance text for composing and formatting conference papers. Please ensure that all template text is removed from your conference paper prior to submission to the conference. Failure to remove the template text from your paper may result in your paper not being published.
}
\end{document}